\documentclass{article}

\usepackage{microtype}
\usepackage{graphicx}
\usepackage{amsmath,amsfonts,amssymb}
\usepackage{subfigure}
\usepackage{booktabs} 

\usepackage{hyperref}

\newcommand{\vv}[1]{\textbf{#1}}

\usepackage[accepted]{icml2021}

\icmltitlerunning{\;}

\begin{document}

\twocolumn[
\icmltitle{Attention-Based Ensemble Pooling for \\ Time Series Forecasting}



\icmlsetsymbol{equal}{*}

\begin{icmlauthorlist}
\icmlauthor{Dhruvit Patel}{umd}
\icmlauthor{Alexander Wikner}{umd}
\end{icmlauthorlist}

\icmlaffiliation{umd}{Department of Physics, University of Maryland, College Park, United States}

\icmlcorrespondingauthor{Dhruvit Patel}{dpp94@umd.edu}
\icmlcorrespondingauthor{Alexander Wikner}{awikner1@umd.edu}

\icmlkeywords{Machine Learning, ICML}

\vskip 0.3in
]



\printAffiliationsAndNotice{\icmlEqualContribution} 

\begin{abstract}
A common technique to reduce model bias in time-series forecasting is to use an ensemble of predictive models and pool their output into an ensemble forecast. In cases where each predictive model has different biases, however, it is not always clear exactly how each model forecast should be weighed during this pooling. We propose a method for pooling that performs a weighted average over candidate model forecasts, where the weights are learned by an attention-based ensemble pooling model. We test this method on two time-series forecasting problems: multi-step forecasting of the dynamics of the non-stationary Lorenz `63 equation, and one-step forecasting of the weekly incident deaths due to COVID-19. We find that while our model achieves excellent valid times when forecasting the non-stationary Lorenz `63 equation, it does not consistently perform better than the existing ensemble pooling when forecasting COVID-19 weekly incident deaths.
\end{abstract}

\section{Introduction and Related Work}
The problem of forecasting the future time evolution of a system is at the heart of many disciplines. Practitioners attempting to forecast a system's evolution typically use a knowledge-based model, a machine-learning-based or statistical model, or a model which combines the two. The use of any model comes with a set of uncertainties and inductive biases associated with that model, e.g., due to how the internal parameters of the model are prescribed and/or constrained, and this may introduce a bias and/or variability into the model forecast. One commonly employed method to mitigate this is to initialize an ensemble of candidate forecasting models and then to aggregate their individual forecasts into a single ensemble forecast, e.g., by using a (un-)weighted average. Ensemble methods have been shown to improve predictive performance in a variety of tasks \cite{sagi,dong,zhou}. However, there remains an open problem, especially when dealing with ensemble learning for forecasting time series: what is the optimal way to pool the individual forecasts of the candidate models into a single ensemble forecast?

A large amount of literature exists discussing methods for ensemble learning. We focus on the specific task of using ensemble learning for time series forecasting. Various previous works have focused on constructing an ensemble pooling module which combines the forecasts of candidate models using either a simple or weighted average \cite{In,pablo,vai,gastinger,wu}. In the cases where a weighted average is used, the weights are determined either by using a statistical model \cite{vai} or by a trainable module \cite{pablo}. In either case, the weights assigned to the different members are typically fixed during the forecasting phase. From the perspective of time series forecasting, where the underlying system evolution may be changing with time in a way such that the different members of the ensemble may perform better or worse depending on the local system dynamics, this is highly disadvantageous. It is desirable to have an ensemble pooling module which can, at every step of the forecast, assign a weight to each candidate model depending on how well that model has performed in the past when similar dynamics were observed. Such an approach would allow us to best leverage the beneficial aspects of the different candidate models.

We propose an attention-based ensemble pooling method which adaptively assigns weights to the different candidate models at each step of the forecast, and we apply this method to two test cases: (1) forecasting the $3$-dimensional Lorenz system, and (2) forecasting the weekly incident deaths during the COVID-19 pandemic. 

In addition, we emphasize that we are considering the situation in which we have access to an ensemble of candidate models for forecasting, but we do not have the ability to re-tune or update the candidate models themselves. This excludes ensemble learning methods such as multi-expert learning \cite{yang}, which rely on jointly training the individual ensemble members and the ensemble pooling network.

\section{Methods}
We propose an ensemble pooling method that performs a weighted average of $M$ candidate models where the weights are chosen adaptively by an attention mechanism based on a representation of the current state of the system (the query) and the set of representations of the candidate models' forecasts (the keys and the values). Below we describe the single-head and multi-head additive attention mechanisms we use in the experiments of this paper. We then describe two modes of operation for our method: (1) the ``open-loop" configuration, and (2) the "closed-loop" configuration.
\subsection{Single-head Additive Attention}
The scalar additive attention weight at time index $j$ for $j=1,...,T$ between the query $\vv{q}_j \in \mathbb{R}^q$ and the $i^{th}$ key $\vv{k}_j^{(i)} \in \mathbb{R}^k$ (in our case, referring to the representation of the forecast of the $i^{th}$ of $M$ candidate models in the ensemble) is given by
\begin{align}\label{eqn:att_weights}
    a_{i,j} = \text{softmax}\Big[\vv{w}_v^\top \tanh(\vv{W}_q\vv{q}_j + \vv{W}_k\vv{k}_j^{(i)} + \vv{b})\Big],
\end{align}
where $\vv{W}_q \in \mathbb{R}^{h \times q}$, $\vv{W}_k \in \mathbb{R}^{h \times k}$, $\vv{w}_v \in \mathbb{R}^h$ and $\vv{b} \in \mathbb{R}^h$ are trainable parameters, and $h$ is the number of hidden units. Our single-head additive attention model ensemble forecast at time index $j$, $\hat{\vv{y}}_j \in \mathbb{R}^d$, is then formed as linear combination of the candidate model forecasts weighted by their respective attention weights,
\begin{align}\label{eqn:sh_pred}
    \hat{\vv{y}}_j = \sum_{i=1}^{M} a_{i,j} F_j^{(i)},
\end{align}
where $F_j^{(i)}$ is the $i^{th}$-model's forecast for time index $j$ given the system state at time index $j-1$.
\subsection{Multi-head Additive Attention}
In many cases, we may want to use a single set of queries, keys, and values to learn multiple behaviors of the target system or to predict multiple quantities simultaneously. In the multi-head attention scheme, we consider $P$ copies of the above described single-head attention block and the set of their corresponding forecasts $\hat{\vv{y}}^{(i)}_j$ for $i=1,...,P$ at time index $j$. The final forecast is obtained by a linear transformation of the concatenation of forecasts from the $P$ heads,
\begin{align}\label{eqn:mh_pred}
    \hat{\vv{y}}_j = \vv{W}_0 \big[\hat{\vv{y}}^{(1)}_j, ..., \hat{\vv{y}}^{(P)}_j \big],
\end{align}
where $\vv{W}_0 \in \mathbb{R}^{d \times (d\times P)}$ is a matrix of trainable parameters.
\subsection{Time-Delayed Embedding of the Queries and Keys}
Since we consider the task of time series forecasting, which has an inherent sequential structure, we formulate the queries and keys by using time-delayed embedding to provide context of the most recent past dynamics. In particular, for a time-delay embedding with $l$ steps into the past, we formulate the queries and keys at time $j$ as the following concatenations:
\begin{align}\label{eqn:q_k_tde}
\begin{split}
    \vv{q}'_j = \big[\vv{q}_j, \vv{q}_{j-1}, ..., \vv{q}_{j-l} \big] \\
    \vv{k}'^{(i)}_j = \big[\vv{k}^{(i)}_j, \vv{k}^{(i)}_{j-1}, ..., \vv{k}^{(i)}_{j-l} \big].
    \end{split}
\end{align}
We note that the time-delayed embedding is meant to provide memory of recent dynamics and can be replaced by a different method that can provide a similar context of past dynamics, e.g., embedding via a recurrent neural network.
\subsection{Modes of Operation}
We utilize our ensemble forecast method in two different modes of operation: (1) the ``open-loop", and (2) the ``closed-loop". In the open-loop configuration, the ground truth data at time index $j-1$ is known and used to obtain the queries $\vv{q}_j$ and the values $F_j^{(i)}$ at time index $j$. The ensemble forecast is then made according to Eq. \ref{eqn:sh_pred} (Eq. \ref{eqn:mh_pred}) for single-head (multi-head) attention. This mode of operation requires that the ``ground truth'' data be available for the time period over which this mode is operated. In the closed-loop configuration, the queries $\vv{q}_j$ are formulated and the values $F_j^{(i)}$ are obtained using the ensemble model forecast at the previous time step $\vv{y}_{j-1}$. In other words, the one-step forecasts of the ensemble model at time index $j-i$ are recycled as inputs to the model at time index $j$. This results in autonomous operation of the ensemble model; thus, operation of this mode only requires access to the ground truth data at initialization. We refer to forecasts made in this configuration as ``multi-step'' forecasts.

Training is performed in the open-loop configuration (i.e., we train our ensemble method on one-step forward forecasts), while testing is performed in either the open-loop or closed-loop configuration, depending on the task.

\section{Results}
\label{sec:results}
\subsection{Non-Stationary Lorenz `63 Equations}\label{lorenz-results}
\subsubsection{Description of True Equations and Ensemble of Candidate Models}\label{sec:lorenz-desc}
As a toy system to demonstrate the potential of our proposed attention-based ensemble pooling model, we consider the Lorenz `63 equations~\cite{lorenz_deterministic_1963}: 
\begin{align}\label{eq:lorenz}
\begin{split}
\frac{du_1(t)}{dt} &= \sigma(u_2(t)-u_1(t)), \\
\frac{du_2(t)}{dt} &= u_1(t)(\rho(t) - u_3(t)) - u_2(t),\\
\frac{du_3(t)}{dt} &= u_1(t)u_2(t) - \beta u_3(t).
\end{split}
\end{align}
The Lorenz `63 equations are a simplified 2D model of atmospheric convection. The system state variables $u_1(t), u_2(t)$, and $u_3(t)$ (which we can represent as the vector $\vv{u}(t) = [u_1(t), u_2(t), u_3(t)]$) represent the rate of convection, the horizontal temperature variation, the vertical temperature variation, respectively, while the model parameters $\sigma, \rho(t),$ and $\beta$ represent the Prandtl number, Rayleigh number, and the size of the system, respectively~\cite{sparrow_lorenz_1982}. For the standard stationary choice of model parameter values ($\sigma = 10$, $\rho(t) = 28$, and $\beta = 8/3$), the dynamics of the Lorenz equations are chaotic. We integrate all sets of Lorenz equations using a $4^{th}$-Order Runge-Kutta integration scheme~\cite{press_numerical_2007} with an integration time step of $0.01$. We choose a trajectory sampling time step of $\Delta t = 0.1$, meaning that we integrate the equations $10$ times before recording the output to form our time series data.

\begin{figure}[t]
    \centering
    \includegraphics[width=0.9\linewidth]{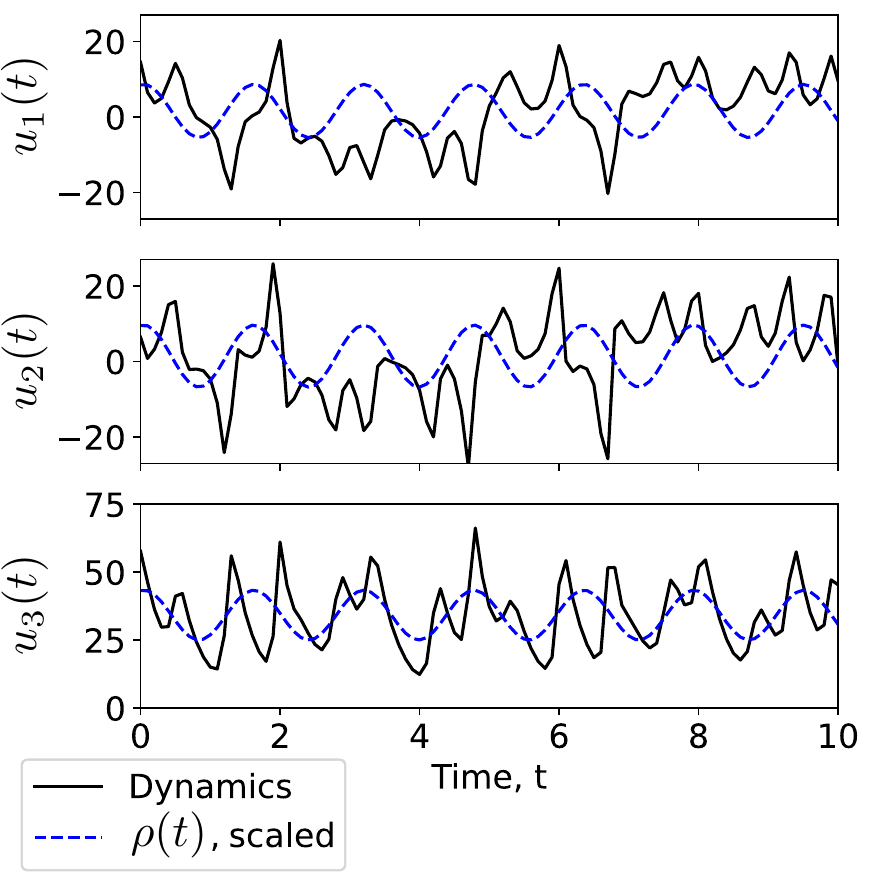}
    \caption{An example trajectory of the Lorenz equations with a non-stationary $\rho(t)$ prescribed by Eq.~\ref{eq:true_rho}. The $\rho(t)$ values are scaled so that they may be referenced against the dynamics.}
    \label{fig:lorenz_dynamics}
\end{figure}

For the true non-stationary dynamics which we will predict using our ensemble pooling method, we set $\sigma$ and $\beta$ to their standard values but allow $\rho(t)$ to vary sinusoidally as follows:
\begin{align}\label{eq:true_rho}
\rho(t)_{True} = 38 - 10\cos(2\pi t/T).
\end{align}
Here, $T$ is the oscillation period. For our tests, we choose an oscillation period of $T = 1.577132$. This period was chosen so that $\rho(t)$ oscillates on a similar time scale to the Lorenz state variables and so that $T$ is not evenly divisible by our chosen sampling time step. This is done so that when we evenly sample initial conditions from a long trajectory for validation, we do not always sample initial conditions obtained with the same initial value of $\rho(t)$. Figure~\ref{fig:lorenz_dynamics} shows an example trajectory of the Lorenz equations with this non-stationary $\rho(t)$. We note that the true dynamics remain chaotic throughout this parameter oscillation.

For our ensemble of candidate models, we use an ensemble of stationary Lorenz equations with the same standard values of $\sigma$ and $\beta$ and the following $M = 11$ evenly-spaced values of $\rho(t)$:
\begin{align}\label{eq:ensemble-rho}
\{\rho(t)_{m, ens}\} = \{28, 30, \dots, 46, 48\}.
\end{align}
\subsubsection{Models Tested, Training, and Evaluation}\label{sec:lorenz-training}
For the attention-based ensemble pooling models tested, we form the queries and keys using the previous true system states and the previous difference between the candidate model one-step forecasts and these true states:
\begin{align}\label{eq:lorenz_keys_queries}
\begin{split}
\vv{q}_j &= \vv{u}_{j-1},\\
\vv{k}_j &= [F_{j-1}^{(1)} - \vv{u}_{j-1}, \dots, F_{j-1}^{(M)}-\vv{u}_{j-1}].
\end{split}
\end{align}
Here, $\vv{u}_j = \vv{u}(j\Delta t)$. For cases when $l > 1$, we form the time-delay embedded queries and keys as described in Eq.~\ref{eqn:q_k_tde}.

We consider the multi-step forecast task for the non-stationary Lorenz equations using the following ensemble pooling techniques:
\begin{enumerate}
    \item Linear regression on the candidate model one-step forecasts (i.e., $\hat{\vv{y}}_{j} = \vv{W}[F_j^{(1)},\dots, F_j^{(M)}]+ \vv{b}$).
    \item Our additive attention ensemble pooling model with an approximately fixed number of trainable parameters. (The size of the hidden dimension, $h$, is inversely scaled by the number of time delay inputs, $l$). We test this method using the following configurations:
    \begin{enumerate}
    \item The multi-step forecast where the model receives a time delay input of length $l = 1, 2, 3, 4, 5$ or $6$ and where the attention weights are re-computed after each one-step forecast (referred to as ``Additive attention'').
    \item The multi-step forecast from the component model which receives the largest attention weight at the beginning of the forecast (referred to as ``Best initial model'').
    \item The multi-step forecast where the attention weights are fixed to their values at the beginning of the forecast (referred to as ``Fixed attention'').
    \end{enumerate}
\end{enumerate}
We also compare with a forecast from a single-hidden-layer feed-forward neural network (FF NN) with a $\tanh$ activation function and approximately the same number of trainable parameters as used in our additive attention model. As in our additive attention ensemble pooling model, we denote the size of its hidden dimension as $h$. This model receives only the time delay vector of previous system states ($\vv{q}'_j$ in Eq.~\ref{eqn:q_k_tde} using $\vv{q}_j$ from Eq.~\ref{eq:lorenz_keys_queries}) as input.

All ensemble pooling models and the FF NN are trained to minimize the mean-squared error between the forecast $\hat{\vv{y}}_j$ and the true state $\vv{y}_j = \vv{u}_j$. The training data is a single trajectory of the non-stationary Lorenz equations of duration $t_{train} = 400$. Given our sampling time step of $\Delta t = 0.1$, our training data sequence thus contains $4000$ samples. Each of the ensemble pooling models and the FF NN are trained using an Adam optimizer~\cite{kingma_adam_2017} with a learning rate of $10^{-3}$. The ensemble pooling methods are trained over $500$ epochs, while the FF NN is trained over $800$ epochs. We validate our forecasts on $200$ validation trajectories evenly sampled from the full validation trajectory of duration $t_{val} = 2560$. Each validation trajectory thus has a duration of $t_{val,i} = 12.8$ and contains $128$ samples.

\begin{figure}[t]
    \centering
    \includegraphics[width=\linewidth]{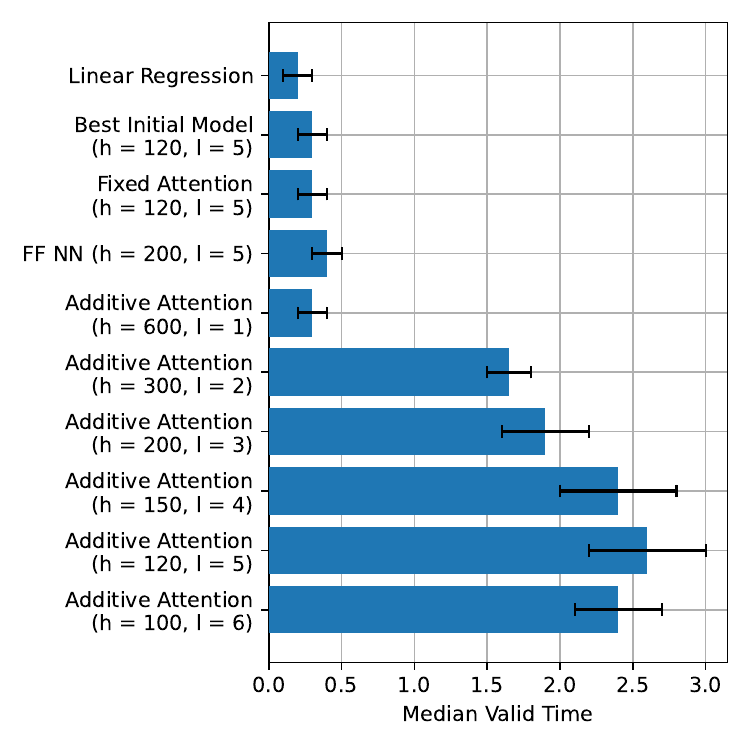}
    \caption{The median valid time of forecasts obtained from each of the ensemble pooling methods tested and from the FF NN. The black error bars on each bar show the maximum of the upper and lower 95\% confidence bounds for the median~\cite{conover_practical_1999} computed over all $200$ validation trajectories.}
    \label{fig:lorenz_vt}
\end{figure}

\subsubsection{forecast Valid Time}\label{sec:lorenz_pred_results}

We evaluate our ensemble forecast performance on the validation trajectories using the valid time of the multi-step forecast, defined as the length of time from the start of the forecast to the time when the mean-squared error in the forecast first exceeds a chosen threshold:
\begin{align}\label{eq:valid_time}
VT = \max_{j\geq0} \Big\{ j\Delta t \:\big|\:\lVert \hat{\vv{y}}_j - \vv{y}_j\rVert^2_2 < \epsilon_{VT} \Big\}.
\end{align}
The mean-squared error between true trajectories with different initial conditions saturates at approximately $200$. We choose an error threshold of $\epsilon_{VT} = 40$.

Figure~\ref{fig:lorenz_vt} shows the median valid time computed over all 200 validation trajectories. Our linear regression ensemble pooling model and our FF NN predictive model do not perform well, only achieving a median valid time of $0.20$ and $0.40$, respectively. Our additive attention ensemble pooling model performs about as poorly as the other methods when the model receives only the most recent model forecast errors and system state as input ($l=1$). However, when a longer time delay input is used ($l>1$), our additive attention model performs substantially better than all other methods, improving from a median valid time of $0.30$ for $l=1$ to $2.60$ for $l=5$ (though we do not see any improvement from $l=5$ to $l=6$ for the particular total number of trainable weights used). We only see this substantial performance increase when the multi-step forecast re-computes the attention weights at each forecast step. The corresponding best initial model and fixed attention model do not improve at all from $l=1$ to $l=5$, and the $l=5$ forecast performs only as well as the additive attention model with $l=1$. These results indicate that our additive attention ensemble pooling model's ability to change which component models to use during a multi-step forecast is key to obtaining good forecast performance. In addition, we see that sufficient knowledge of past states and component model error is needed in order for our ensemble pooling model to perform well.

\subsubsection{Attention Weights during Multi-Step forecast}
\begin{figure}[t]
    \centering
    \includegraphics[width=\linewidth]{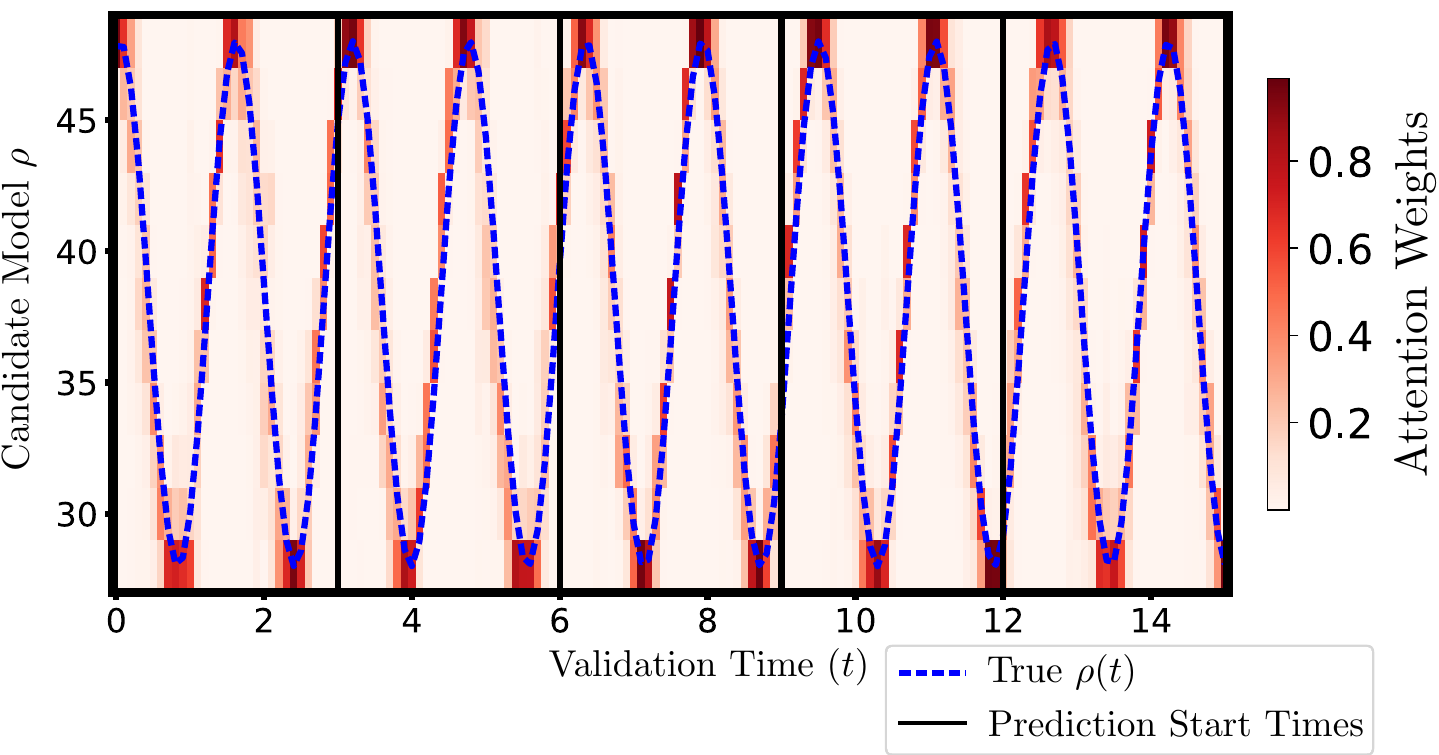}
    \caption{The attention weights assigned to each of the candidate models by our additive attention ensemble pooling model with $h=120$ and $l=5$. Each vertical black line (shown at $t = 0, 3, 6, 9,$ and $12$) marks where the forecast is re-initialized using the true validation data. The model then performs a multi-step forecast until the next re-initialization. The weight of each candidate model is given by the color axis, while the dashed blue line shows the true value of $\rho(t)$.}
    \label{fig:lorenz_weights}
\end{figure}
Figure~\ref{fig:lorenz_weights} shows the attention weights assigned to each of the candidate models with different stationary $\rho(t)$ values during $5$ multi-step forecasts of a validation trajectory, each of duration $3$, made by our additive attention ensemble pooling model with $h=120$ and $l = 5$. We see that during these forecasts, the true $\rho(t)$ value and the $\rho(t)$ value of the candidate model assigned the highest weight are very similar. This indicates that even during a multi-step forecast where, after $l$ steps, the model no longer receives any inputs from the true data, our model is still able to anticipate the correct change in $\rho(t)$ from the past one-step forecasts.

\subsection{Weekly Incident Deaths from Covid-19}\label{sec:covid-results}
\subsubsection{Description of the COVID-19 data set and the evaluation metric}\label{sec:covid-desc}
In order to test our method on a real-world data set, we consider the task of predicting the weekly incident number of deaths ($\Delta t =$ 1 week) during the COVID-19 pandemic using data provided by the COVID-19 Forecast Hub \cite{covidhub}, founded in March 2020 by the Reich Lab of the University of Massachusetts. The COVID-19 Forecast Hub maintains an up-to-date record for forecasts of COVID-19 cases, hospitalizations, and deaths in the US, created by dozens of leading infectious disease modeling teams around the world, in coordination with the US CDC. Participating teams submit a probabilistic forecast for future days, weeks, and months for the county, state and national levels in the US. Each forecast consists of a set of quantiles of the predictive distribution at prescribed levels (e.g., 10\%, 20\%, ... 95\%, 98\%). In addition to providing the forecasts of the different participating teams, the COVID-19 Forecast Hub also provides an ensemble forecast which it generates by aggregating the forecasts of the different participating teams. While the method of ensemble pooling used by the COVID-19 Forecast Hub has varied over the course of the pandemic, the most current method performs a weighted ensemble average of the models with the best performance, as measured by their Weighted Interval Score (WIS) \cite{wis}, in the 12 weeks prior to the forecast date. The component models are assigned weights that are a function of their relative WIS during those 12 weeks. 

The WIS \cite{covidhub_wis,wis} is a proper scoring rule which assesses the quality of a probabilistic forecast (described by quantiles of the predictive cumulative distribution $P$) based on the predictive cumulative distribution and on the observed event. For an observed event $y$, and for the quantile forecast at level $\alpha$, the Interval Score (IS) is given by:
\begin{multline}\label{eqn:IS}
    IS_\alpha (P,y) = (u-l) + 2/\alpha \times (l-y) \times \vv{1}(y < l) + \\
    2/\alpha \times (y-u) \times \vv{1}(y > u),
\end{multline}
where \vv{1} is the indicator function which equals $1$ if its argument is true and $0$ otherwise. The terms $l$ and $u$ denote the $\alpha/2$ and $1 - \alpha/2$ quantiles of $P$. The WIS is calculated for a probabilistic forecast that provides the forecast at several $\alpha_i$ levels for $i=1, ..., K$, 
\begin{multline}\label{eqn:WIS}
    WIS_{\alpha_0:K} (P,y) = \frac{1}{K+1/2} \times \Bigl( w_0 \times |y-m| + \\
    \sum_{k=1}^K\{w_k \times IS_{\alpha_k} (P,y)\} \Bigr),
\end{multline}
where $w_i$ are the weights, which we set to $w_i = \alpha_i /2$ (and $w_0 = 1/2$). The WIS is a \emph{negatively oriented} score, meaning that the lower the score the higher the quality of the probabilistic forecast, and (for the above choice of weights $w_i$) approximates the Continuous Ranked Probability Score \cite{covidhub_wis,wis_crps}.

\subsubsection{COVID-19 Candidate Model Selection and Data Pre-processing}
A major challenge presented by the COVID-19 data set is that of missing data. While many teams have participated in the COVID-19 forecast challenge hosted by the COVID-19 Forecast Hub since its creation in March $2020$ to the present, most teams have not submitted forecasts for this entire span of time. In addition, many teams have not submitted forecasts consistently week-to-week after their first submission. Due to this, we have chosen as candidate models the $M=9$ models which have submitted forecasts of weekly incident deaths for every US state and DC most consistently from May $2020$ through October $2022$. For any missing forecasts in this period, the weekly incident deaths for a given team at a given location and for a given quantile is filled in by:
\begin{enumerate}
    \item Linear interpolation between weekly forecasts if the gap of missing data spans less than three weeks.
    \item Using the candidate model ensemble mean if the gap of missing data spans more than two weeks, and if other candidate models in the ensemble have made forecasts for this time.
    \item Using the nearest past or future value if the gap of missing data spans more than two weeks, and if no other model in the ensemble has made a forecast for this time.
\end{enumerate}
We use the following $21$ quantiles from each candidate model forecast for every US state and DC:
\begin{align}
\begin{split}
F_j^{(i)} = [&F_{j,\alpha_1}^{(i)}, \dots, F_{j,\alpha_{21}}^{(i)}],\text{ where}\\
\{\alpha\} = \{&0.5, 0.01, 0.025, 0.05, 0.1, 0.15, 0.2, 0.25,\\ & 0.3,0.35, 0.4, 0.45, 0.55, 0.6, 0.65, 0.7, \\&0.75, 0.8, 0.85, 0.9, 0.95, 0.975, 0.99\}.
\end{split}\label{eq:quantiles}
\end{align}
The forecasts made by each of the ensemble pooling models we will test will predict these same quantiles.

\subsubsection{Models Tested, Training, and Evaluation}
For the attention-based ensemble pooling models tested, the queries are formed using the previous true number of weekly incident deaths, which we denote as $u_{j-1}$. The keys are formed using the difference between the candidate models's previous median forecasts ($\alpha_1=0.5$) and the previous number of weekly incident deaths, in addition to all other candidate model forecast quantiles. We therefore have:
\begin{align}\label{eq:lorenz_keys_queries}
\begin{split}
\vv{q}_j = & u_{j-1},\\
\vv{k}_j = &[F_{j-1, \alpha_1}^{(1)} - u_{j-1}, F_{j-1, \alpha_2}^{(1)}, \dots, F_{j-1, \alpha_{21}}^{(1)},\dots,\\
&F_{j-1, \alpha_1}^{(M)} - u_{j-1}, F_{j-1, \alpha_2}^{(M)}, \dots, F_{j-1, \alpha_{21}}^{(M)} ].
\end{split}
\end{align}
For cases when $l > 1$, we form the time-delay embedded queries and keys as described in Eq.~\ref{eqn:q_k_tde}. We note that we also tried using different queries and keys, including adding the number of incident COVID-19 cases to the queries and the previous WIS values for each candidate model forecast to the keys. We found, however, that these additions did not consistently improve performance, so we have chosen to omit them.

Since we only have access to individual model forecasts and not the models themselves, we consider the one-step forecast task for the COVID-19 weekly incident deaths using the following ensemble pooling models:
\begin{enumerate}
\item Linear regression on the $l=5$ most-recent candidate model one-step forecasts:
\begin{align}
\begin{split}\hat{\vv{y}}_{j} = \vv{W}[&F_j^{(1)},\dots, F_{j-(l-1)}^{(1)}, \dots,\\&F_j^{(M)}, \dots, F_{j-(l-1)}^{(M)}]+ \vv{b}.
\end{split}
\end{align}
\item Our additive attention ensemble pooling model with $h = 1000$ and $l = 5$.
\item Our multi-head attention ensemble pooling model with $h = 100$, $l = 5$, and $p = 21$.
\end{enumerate}
We compare the resulting ensemble forecasts with the COVID-19 Forecast Hub ensemble forecast. We note that while the ensemble pooling methods we test use only the $M=9$ candidate models with the most forecast data, the COVID-19 Forecast Hub ensemble forecast may make use of any of the $124$ candidate models for which it has forecasts from at a particular time.

\begin{figure}[t]
    \centering
    \includegraphics[width=\linewidth]{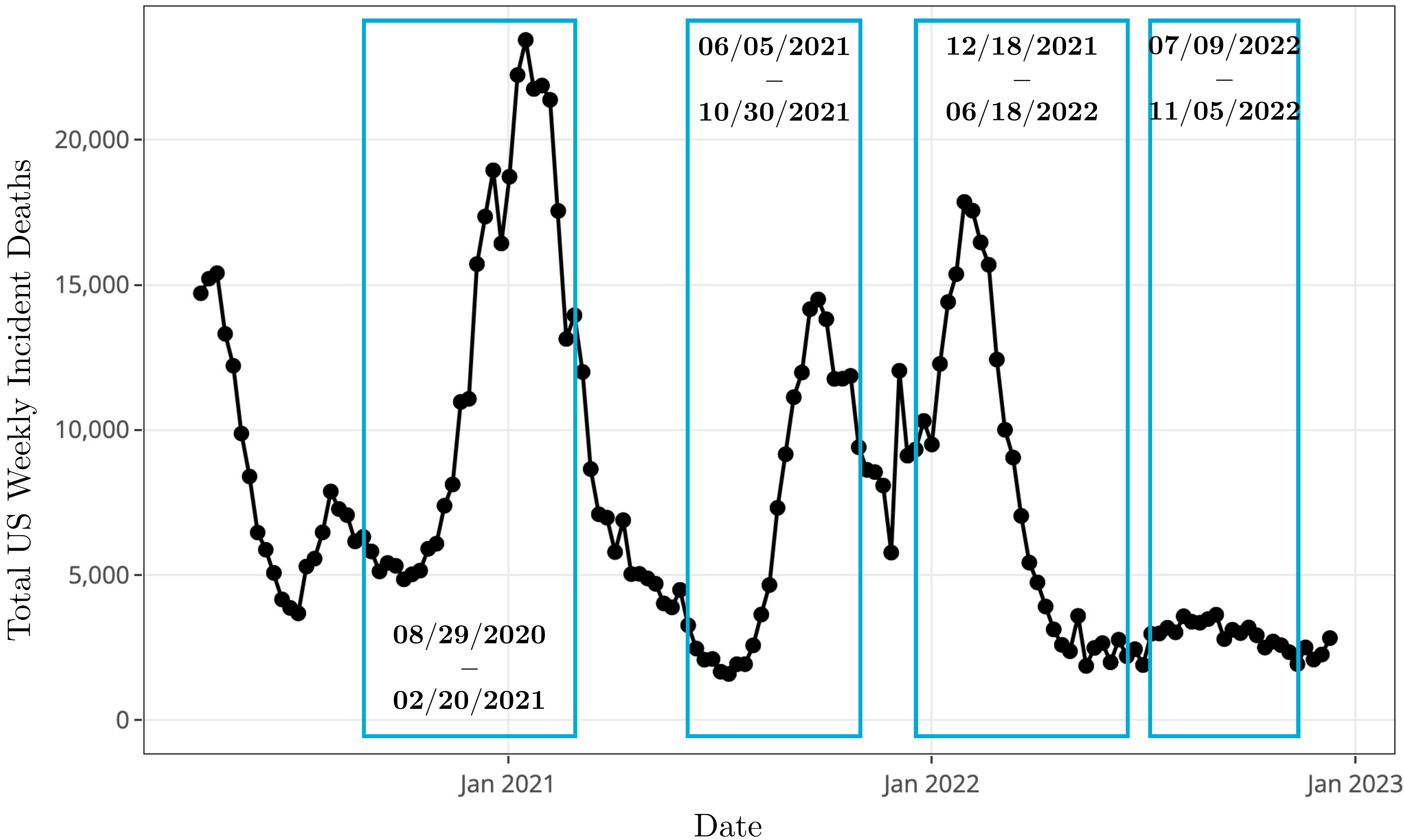}
    \caption{The validation periods used to evaluate the COVID-19 weekly incident death ensemble forecasts. The black line shows the total US weekly incident deaths, and each blue box bounds a particular validation period.}
    \label{fig:covid_test_periods}
\end{figure}
We validate our ensemble forecasts within the 4 periods shown in Fig.~\ref{fig:covid_test_periods} on data from every US state and DC. For each validation period, we train each of our ensemble pooling models using the ground truth and candidate model forecast data from outside of this period from every US state and DC. The models are trained using an Adam optimizer with a learning rate of $10^{-5}$ over $200$ epochs. For the Additive and Multi-head attention models, we use a small weight decay of $10^{-4}$ and $10^{-5}$, respectively.

The loss function for each of our ensemble pooling models is the mean WIS computed using the $\alpha$ values
\begin{align}
\begin{split}
\{\alpha_{WIS}\} = \{&0.02, 0.05, 0.1, 0.3, 0.4,\\ &0.5, 0.6, 0.7, 0.8, 0.9\},
\end{split}
\end{align}
and with the observed event $y = u_j$, the true observed weekly incident deaths. One can see from Eq.~\ref{eqn:WIS} that the $21$ forecast quantiles produced by our models (Eq.~\ref{eq:quantiles}) are sufficient to compute the WIS for these $\alpha$ values. We will evaluate the accuracy of our forecasts during validation using this WIS.
\subsubsection{Results}\label{sec:covid_results}
\begin{center}
\begin{table}
\centering
\resizebox{\columnwidth}{!}{
\begin{tabular}{|c||c|c|c|c|}
\hline
&\multicolumn{4}{|c|}{Mean WIS over All States}\\\hline
\begin{tabular}{c}Validation\\Period\end{tabular}
&\begin{tabular}{c}08/29/2020\\-\\02/20/2021\end{tabular}
&\begin{tabular}{c}06/05/2021\\-\\10/30/2021\end{tabular}
&\begin{tabular}{c}12/21/2021\\-\\06/18/2022\end{tabular}
&\begin{tabular}{c}07/09/2022\\-\\11/05/2022\end{tabular}\\\hline
\hline
\begin{tabular}{c}COVID-19\\Forecast Hub\\Ensemble\end{tabular}&
\vv{35.12} & 22.13 & \vv{32.86} & \vv{12.60}\\\hline
\begin{tabular}{c}Linear\\Regression\end{tabular}&
35.75 & \vv{20.42} & 35.98 & 13.80\\\hline
\begin{tabular}{c}Additive\\Attention\end{tabular}&
38.27 & 20.72 & 35.51 & 12.92\\\hline
\begin{tabular}{c}Multi-head\\Attention\end{tabular}&
36.88 & 20.86 & 34.74 & 12.94\\\hline
\end{tabular}
}
\caption{Mean WIS values computed by averaging forecast WIS over dates in each validation period and over forecasts for all states and DC. Bold text numbers mark the lowest WIS ensemble forecast for that particular validation period.}
\label{table:covid_results}
\end{table}
\end{center}
We show the mean WIS computed over all states and DC for each validation period in Table~\ref{table:covid_results}. We see that while our attention-based models do obtain a smaller WIS than the COVID-19 Forecast Hub ensemble during the summer 2021 validation period, the linear regression ensemble pooling model actually performs better than either of the attention-based models during this period. This improved performance over the attention-based models holds for the first two validation periods, but not for the final two. In these last two periods, the linear regression model has the highest mean WIS. Though our attention-based models do perform better than linear regression in these periods, they are not obtain a smaller WIS than the COVID-19 Forecast Hub ensemble. The same is true for all of the ensemble pooling models we tested in the earliest validation period.
\begin{figure*}[t]
    \centering
    \includegraphics[width=0.8\linewidth]{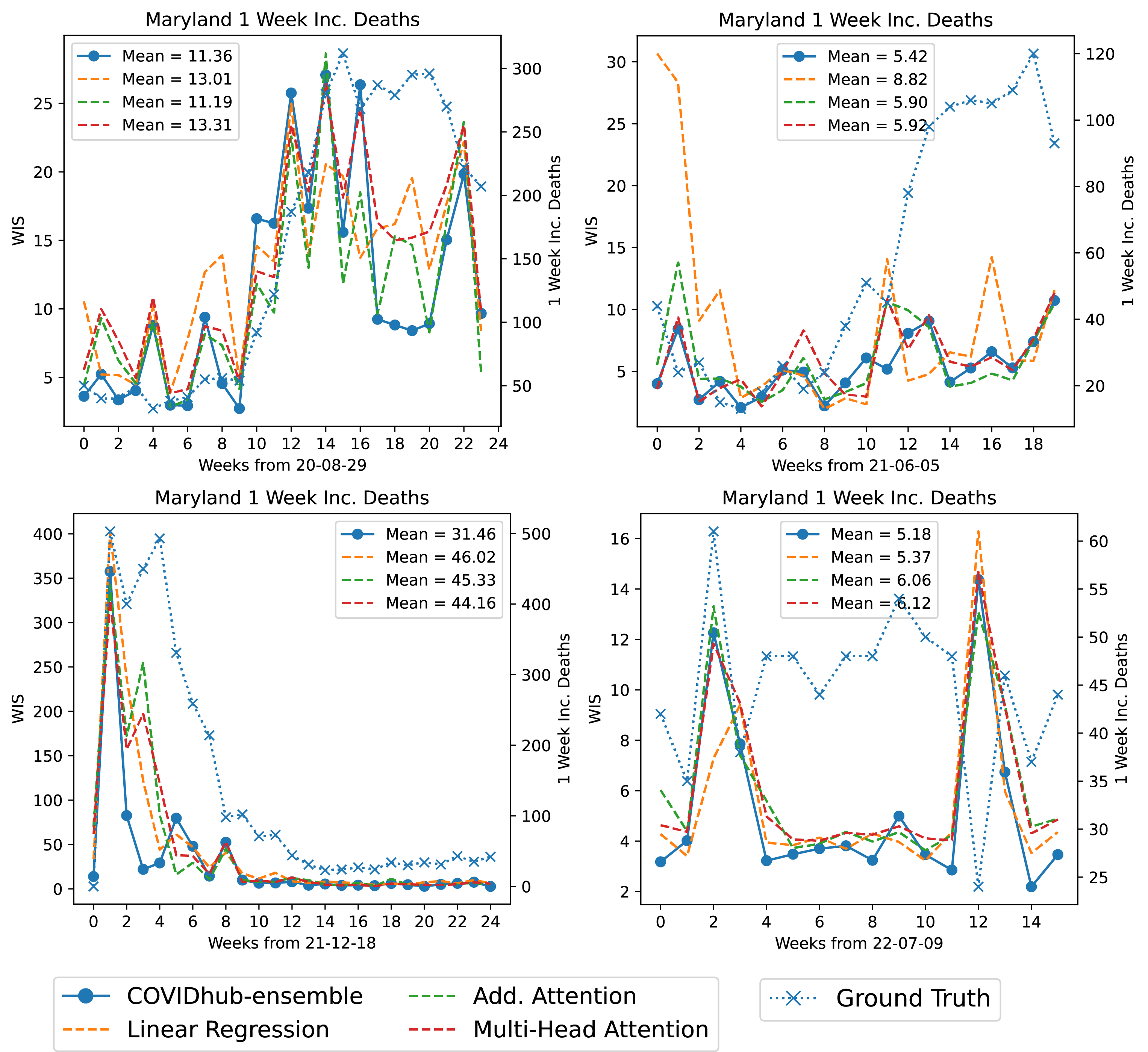}
    \caption{Each panel shows the WIS for forecasts of the weekly incident deaths in Maryland made for each week during each of the four validation periods. Each panel also shows the ground truth number of weekly incident deaths. The scale for the WIS is shown along each right vertical axis, while the scale for the number of weekly incident deaths is shown along each left vertical axis. The target forecast week is shown along each horizontal axis. The legend in each panel shows the mean WIS computed over the corresponding validation period.}
    \label{fig:covid_maryland_results}
\end{figure*}

As an example of our ensemble pooling model performance in a single state, we show the WIS for forecasts of the weekly incident deaths in Maryland over time in each validation period in Fig.~\ref{fig:covid_maryland_results}. The overall performance is similar to that averaged over all states; while the additive attention ensemble pooling model has a lower mean WIS than the other models during the validation period beginning from 08/29/2020, it does not during any of the other periods. In addition, none of the ensemble pooling models that we trained consistently have a lower mean WIS than the other over all of the validation periods. From these panels, one can see that the mean WIS is heavily weighted towards the model's performance during a wave, when the number of weekly incident deaths is high. For example, looking at week 3 in the panel beginning from 12/18/2021, we see that the WIS for each of the ensemble pooling models we trained is substantially higher than the COVID-19 Forecast Hub ensemble. Despite the fact that the models all perform similarly before and after this difference, the COVID-19 Forecast Hub ensemble still has a much smaller mean WIS than the other models due to this short but sharp performance difference.
\section{Conclusion}\label{sec:conclusion}
We propose a method for creating an ensemble forecast for systems where one has access to a set of candidate models and ground truth time series data. Our proposed method combines the candidate model forecasts using a weighted average, where the weights are obtained by training an attention-based machine learning model. We evaluate the performance of this model on two test cases: the dynamics of the non-stationary Lorenz `63 equations, and the weekly incident deaths from COVID-19.

We find that our single-head attention model's ensemble forecasts of the non-stationary Lorenz `63 equations have a substantially longer median valid time than the other ensemble pooling models tested. In addition, our model is able to correctly select the candidate model with the closest internal model parameter to the true system during a multi-step forecast, even after the model no longer receives any input from the ground truth data. 

On the other hand, we find that neither our single-head nor our multi-head attention models consistently improve the mean WIS of one-step forecasts of the weekly incident deaths due to COVID-19 over the COVID-19 Forecast Hub ensemble forecast. There are a number of reasons why this may be the case: the COVID-19 Forecast Hub ensemble uses a far greater number of candidate models than we were able to use in the ensemble pooling models we trained. In addition, the scheme we used to fill in missing candidate model forecast data may have introduced unrealistic biases. The ground truth data may also suffer from erroneous reporting of deaths or from deaths which are reported later than they actually occurred, making it more difficult to learn when particular candidate models should be used. Finally, our attenion-based ensemble pooling model assumes that the biases of the candidate models are not changing, or do not change much, over time; the models on the COVID-19 Forecast Hub, however, have been continuously updated since they began submitting forecasts. It may therefore not be possible to learn these models's biases without adding an explicit time-dependence.

While it may be possible to improve our attention-based ensemble pooling model's performance on the COVID-19 forecast problem by adding additional context to the queries and keys, the inherent challenges in this data set remain. We therefore would like to test this method on a different real-world data set where we have both ample training data and fixed model biases. Namely, we intend to use an ensemble of numerical terrestrial climate models to forecast the El Ni\~{n}o Southern Oscillation.
\section*{Code Availability}
The code that supports the findings of this study may be found at \url{https://github.com/awikner/denpool}.

\bibliography{main}

\begin{thebibliography}{18}
\providecommand{\natexlab}[1]{#1}
\providecommand{\url}[1]{\texttt{#1}}
\expandafter\ifx\csname urlstyle\endcsname\relax
  \providecommand{\doi}[1]{doi: #1}\else
  \providecommand{\doi}{doi: \begingroup \urlstyle{rm}\Url}\fi

\bibitem[Bracher et~al.(2021)Bracher, Ray, Gneiting, and Reich]{covidhub_wis}
Bracher, J., Ray, E.~L., Gneiting, T., and Reich, N.~G.
\newblock Evaluating epidemic forecasts in an interval format.
\newblock \emph{{PLOS} Computational Biology}, 17\penalty0 (2):\penalty0
  e1008618, feb 2021.
\newblock \doi{10.1371/journal.pcbi.1008618}.
\newblock URL \url{https://doi.org/10.1371%2Fjournal.pcbi.1008618}.

\bibitem[Conover(1999)]{conover_practical_1999}
Conover, W.~J.
\newblock \emph{Practical {{Nonparametric Statistics}}}.
\newblock {Wiley}, {New York, NY, USA}, third edition, 1999.
\newblock ISBN 978-0-471-16068-7.

\bibitem[Cramer et~al.(2021)Cramer, Huang, Wang, Ray, Cornell, Bracher,
  Brennen, Castro~Rivadeneira, Gerding, House, Jayawardena, Kanji, Khandelwal,
  Le, Niemi, Stark, Shah, Wattanachit, Zorn, Reich, and Consortium]{covidhub}
Cramer, E.~Y., Huang, Y., Wang, Y., Ray, E.~L., Cornell, M., Bracher, J.,
  Brennen, A., Castro~Rivadeneira, A.~J., Gerding, A., House, K., Jayawardena,
  D., Kanji, A.~H., Khandelwal, A., Le, K., Niemi, J., Stark, A., Shah, A.,
  Wattanachit, N., Zorn, M.~W., Reich, N.~G., and Consortium, U. C.-. F.~H.
\newblock The united states covid-19 forecast hub dataset.
\newblock \emph{medRxiv}, 2021.
\newblock \doi{10.1101/2021.11.04.21265886}.
\newblock URL
  \url{https://www.medrxiv.org/content/10.1101/2021.11.04.21265886v1}.

\bibitem[Dong et~al.(2020)Dong, Yu, Cao, Shi, and Ma]{dong}
Dong, X., Yu, Z., Cao, W., Shi, Y., and Ma, Q.
\newblock A survey on ensemble learning.
\newblock \emph{Frontiers of Computer Science}, 14:\penalty0 241--258, 2020.
\newblock \doi{https://doi.org/10.1007/s11704-019-8208-z}.

\bibitem[Gastinger et~al.(2021)Gastinger, Nicolas, Stepić, Schmidt, and
  Schülke]{gastinger}
Gastinger, J., Nicolas, S., Stepić, D., Schmidt, M., and Schülke, A.
\newblock A study on ensemble learning for time series forecasting and the need
  for meta-learning, 2021.
\newblock URL \url{https://arxiv.org/abs/2104.11475}.

\bibitem[Gneiting \& Raftery(2007)Gneiting and Raftery]{wis}
Gneiting, T. and Raftery, A.~E.
\newblock Strictly proper scoring rules, prediction, and estimation.
\newblock \emph{Journal of the American Statistical Association}, 102\penalty0
  (477):\penalty0 359--378, 2007.
\newblock \doi{10.1198/016214506000001437}.
\newblock URL \url{https://doi.org/10.1198/016214506000001437}.

\bibitem[Gneiting \& Ranjan(2011)Gneiting and Ranjan]{wis_crps}
Gneiting, T. and Ranjan, R.
\newblock Comparing density forecasts using threshold-and quantile-weighted
  scoring rules.
\newblock \emph{Journal of Business \& Economic Statistics}, 29\penalty0
  (3):\penalty0 411--422, 2011.
\newblock ISSN 07350015.
\newblock URL \url{http://www.jstor.org/stable/23243806}.

\bibitem[In \& Jung(2022)In and Jung]{In}
In, Y. and Jung, J.-Y.
\newblock Simple averaging of direct and recursive forecasts via partial
  pooling using machine learning.
\newblock \emph{International Journal of Forecasting}, 38\penalty0
  (4):\penalty0 1386--1399, 2022.
\newblock ISSN 0169-2070.
\newblock \doi{https://doi.org/10.1016/j.ijforecast.2021.11.007}.
\newblock URL
  \url{https://www.sciencedirect.com/science/article/pii/S0169207021001813}.
\newblock Special Issue: M5 competition.

\bibitem[Kingma \& Ba(2017)Kingma and Ba]{kingma_adam_2017}
Kingma, D.~P. and Ba, J.
\newblock Adam: {{A Method}} for {{Stochastic Optimization}}, January 2017.

\bibitem[Lorenz(1963)]{lorenz_deterministic_1963}
Lorenz, E.~N.
\newblock Deterministic {{Nonperiodic Flow}}.
\newblock \emph{Journal of Atmospheric Sciences}, 20\penalty0 (2):\penalty0
  130--141, March 1963.
\newblock ISSN 0022-4928, 1520-0469.
\newblock \doi{10.1175/1520-0469(1963)020<0130:DNF>2.0.CO;2}.

\bibitem[Montero-Manso et~al.(2020)Montero-Manso, Athanasopoulos, Hyndman, and
  Talagala]{pablo}
Montero-Manso, P., Athanasopoulos, G., Hyndman, R.~J., and Talagala, T.~S.
\newblock Fforma: Feature-based forecast model averaging.
\newblock \emph{International Journal of Forecasting}, 36\penalty0
  (1):\penalty0 86--92, 2020.
\newblock ISSN 0169-2070.
\newblock \doi{https://doi.org/10.1016/j.ijforecast.2019.02.011}.
\newblock URL
  \url{https://www.sciencedirect.com/science/article/pii/S0169207019300895}.
\newblock M4 Competition.

\bibitem[Press(2007)]{press_numerical_2007}
Press, W.~H.
\newblock \emph{Numerical {{Recipes}}: {{The Art}} of {{Scientific
  Computing}}}.
\newblock {Cambridge University Press}, 2007.
\newblock ISBN 978-978-052-188-2.

\bibitem[Sagi \& Rokach(2018)Sagi and Rokach]{sagi}
Sagi, O. and Rokach, L.
\newblock Ensemble learning: A survey.
\newblock \emph{WIREs Data Mining and Knowledge Discovery}, 8\penalty0
  (4):\penalty0 e1249, 2018.
\newblock \doi{https://doi.org/10.1002/widm.1249}.
\newblock URL
  \url{https://wires.onlinelibrary.wiley.com/doi/abs/10.1002/widm.1249}.

\bibitem[Sparrow(1982)]{sparrow_lorenz_1982}
Sparrow, C.
\newblock \emph{The {{Lorenz Equations}}: {{Bifurcations}}, {{Chaos}}, and
  {{Strange Attractors}}}, volume~41 of \emph{Applied {{Mathematical
  Sciences}}}.
\newblock {Springer}, {New York, NY}, 1982.
\newblock ISBN 978-0-387-90775-8 978-1-4612-5767-7.
\newblock \doi{10.1007/978-1-4612-5767-7}.

\bibitem[Vaiciukynas et~al.(2021)Vaiciukynas, Danenas, Kontrimas, and
  Butleris]{vai}
Vaiciukynas, E., Danenas, P., Kontrimas, V., and Butleris, R.
\newblock Two-step meta-learning for time-series forecasting ensemble.
\newblock \emph{IEEE Access}, 9:\penalty0 62687--62696, 2021.
\newblock \doi{10.1109/ACCESS.2021.3074891}.

\bibitem[Wu \& Levinson(2021)Wu and Levinson]{wu}
Wu, H. and Levinson, D.
\newblock The ensemble approach to forecasting: A review and synthesis.
\newblock \emph{Transportation Research Part C: Emerging Technologies},
  132:\penalty0 103357, 2021.
\newblock ISSN 0968-090X.
\newblock \doi{https://doi.org/10.1016/j.trc.2021.103357}.
\newblock URL
  \url{https://www.sciencedirect.com/science/article/pii/S0968090X21003594}.

\bibitem[Yang et~al.(2020)Yang, Yuan, Zhu, Yu, and Li]{yang}
Yang, C., Yuan, K., Zhu, Q., Yu, W., and Li, Z.
\newblock Multi-expert learning of adaptive legged locomotion.
\newblock \emph{Science Robotics}, 5\penalty0 (49), dec 2020.
\newblock \doi{10.1126/scirobotics.abb2174}.
\newblock URL \url{https://doi.org/10.1126%2Fscirobotics.abb2174}.

\bibitem[Zhou(2021)]{zhou}
Zhou, Z.-H.
\newblock \emph{Ensemble Learning}, pp.\  181--210.
\newblock Springer Singapore, Singapore, 2021.
\newblock ISBN 978-981-15-1967-3.
\newblock \doi{10.1007/978-981-15-1967-3_8}.
\newblock URL \url{https://doi.org/10.1007/978-981-15-1967-3_8}.

\end{thebibliography}
\bibliographystyle{icml2021}

\end{document}